 \documentclass[pmlr,twocolumn,10pt]{jmlr} % W&CP article

\usepackage{multirow, makecell}
\usepackage{booktabs}
\usepackage{multicol}
\usepackage{dblfloatfix}

\usepackage{tablefootnote}
\usepackage{array}
\newcolumntype{L}[1]{>{\raggedright\let\newline\\\arraybackslash\hspace{0pt}}m{#1}}

\usepackage[load-configurations=version-1]{siunitx} % newer version 
\theorembodyfont{\upshape}
\theoremheaderfont{\scshape}
\theorempostheader{:}
\theoremsep{\newline}

\jmlrvolume{LEAVE UNSET}
\jmlryear{2021}
\jmlrsubmitted{LEAVE UNSET}
\jmlrpublished{LEAVE UNSET}
\jmlrworkshop{Machine Learning for Health (ML4H) 2021} % W&CP title

\title[CEHR-BERT]{CEHR-BERT: Incorporating temporal information from structured EHR data to improve prediction tasks}

\author{%
\Name{\footnotesize Chao Pang} \Email{\footnotesize cp3016@cumc.columbia.edu}\\
\Name{\footnotesize Xinzhuo Jiang} \Email{\footnotesize xj2193@cumc.columbia.edu}\\
\Name{\footnotesize Krishna S. Kalluri} \Email{\footnotesize kk3326@cumc.columbia.edu}\\
\Name{\footnotesize Matthew Spotnitz} \Email{\footnotesize mes2165@cumc.columbia.edu}\\
\addr \footnotesize Columbia University Irving Medical Center
\AND
\Name{\footnotesize RuiJun Chen} \Email{\footnotesize rchen@geisinger.edu}\\
\addr \footnotesize Geisinger
\AND
\Name{\footnotesize Adler Perotte} \Email{\footnotesize ajp2120@cumc.columbia.edu}\\
\Name{\footnotesize Karthik Natarajan} \Email{\footnotesize kn2174@cumc.columbia.edu}\\
\addr \footnotesize Columbia University Irving Medical Center
}

\begin{document}

\maketitle

\begin{abstract}
Embedding algorithms are increasingly used to represent clinical concepts in healthcare for improving machine learning tasks such as clinical phenotyping and disease prediction. Recent studies have adapted state-of-the-art bidirectional encoder representations from transformers (BERT) architecture to structured electronic health records (EHR) data for the generation of contextualized concept embeddings, yet do not fully incorporate temporal data across multiple clinical domains. Therefore we developed a new BERT adaptation, CEHR-BERT, to incorporate temporal information using a hybrid approach by augmenting the input to BERT using artificial time tokens, incorporating time, age, and concept embeddings, and introducing a new second learning objective for visit type. CEHR-BERT was trained on a subset of  clinical data from Columbia University Irving Medical Center-York Presbyterian Hospital, which includes 2.4M patients, spanning over three decades, and tested using 4-fold evaluation on the following prediction tasks: hospitalization, death, new heart failure (HF) diagnosis, and HF readmission. Our experiments show that CEHR-BERT outperformed existing state-of-the-art clinical BERT adaptations and baseline models across all 4 prediction tasks in both ROC-AUC and PR-AUC. CEHR-BERT also demonstrated strong few-shot learning capability, as our model trained on only 5\% of data outperformed comparison models trained on the entire data set. Ablation studies to better understand the contribution of each time component showed incremental gains with every element, suggesting that CEHR-BERT’s incorporation of artificial time tokens,  time/age embeddings with concept embeddings, and the addition of the second learning objective represents a promising approach for future BERT-based clinical embeddings.
\end{abstract}

\section{Introduction}
\label{sec:intro}

Embedding algorithms, widely used for obtaining low dimensional vector representations of words in natural language processing (NLP) applications, have been increasingly adapted for the representation of clinical concepts in healthcare to improve the development of clinical phenotypes and disease prediction  \citep{Glicksberg2018}. Recent advances in contextualized representations such as Bidirectional encoder representations from transformers (BERT) have revolutionized the NLP field, achieving state of the art performance on all benchmark tasks \citep{Devlin2019, Peters2018}. However, there have been few efforts to apply BERT to structured electronic health record (EHR) data for generating contextualized concept embeddings despite promising results from early BERT adaptations demonstrating improved performance compared to classic embedding algorithms such as word2vec and GloVe  \citep{Rasmy2021, Li2020}.

Despite the differences between structured EHR and text data, a common practice used in the aforementioned BERT adaptations and other classic embedding algorithms is to treat a patient’s medical history as a text document where medical concepts are treated as words and ordered chronologically  \citep{Beam2020, Choi2016, Xiang2019}. Although  this  representation  could  capture the rich contextual information of a patient’s medical history, the  temporal  intervals  between  medical concepts or visits are not preserved; as a consequence, BERT models trained using this patient representation cannot fully leverage temporal information, limiting their performance in downstream prediction tasks. Another challenge is that BERT’s second learning objective – Next Sentence Prediction (NSP) \citep{Devlin2019} does not apply in the context of structured EHR data as the entire patient history is treated as a single sentence. A common approach adopted by others is to disable NSP and pre-train BERT only with Masked Language Modeling (MLM) \citep{Li2020}, yet there is abundant information in structured EHR data that could be leveraged for designing a new secondary learning objective to improve BERT’s performance in downstream prediction tasks.

In this paper, we focus on adapting the original BERT architecture for structured EHR data in order to improve disease predictions. We propose a new BERT architecture called CEHR-BERT, where we combine two approaches for encoding temporal information of the structured EHR data by:
1) modifying the patient sequence representation through the insertion of artificial tokens between visits to indicate the time intervals; 2) concatenating both age embeddings and time embeddings to concept embeddings to form temporal concept embeddings. Additionally, we designed a second learning objective -- Visit Type Prediction (VTP) for CEHR-BERT that leverages heterogeneous EHR data to further boost the performance of BERT.

\section{Related work}
\label{sec:related_work}
A small number of recent studies have sought to adapt BERT for structured EHR data and demonstrated significant performance improvements in their respective evaluations. However, these studies were often limited to a single clinical domain, single visit, or limited in their consideration of time, without fully utilizing the richness of a patient's full medical history.

\citet{Li2020} described the first BERT adaptation for structured EHR data named  BEHRT, which pioneered the idea of utilizing multiple types of embeddings to represent patient history, including concept embeddings, visit segment embeddings, age embeddings, and positional embeddings. In addition, the authors inserted  \textbf{SEP} tokens between visits to indicate the boundaries of visits and enable the execution of BERT as-is on structured EHR data without any modification of the encoder. However, this study only included diagnosis codes in a patient sequence and excluded other clinical domains such as procedures and medications that contain valuable contextual information. In addition, their evaluation focused on diagnosis code prediction instead of disease prediction based on phenotypes.

Another BERT adaptation named G-BERT \citep{Shang2019} extended BERT to incorporate a graph neural network (GNN). The key idea was to leverage prior knowledge from well crafted medical ontologies to guide the learning of concept embeddings. However, G-BERT was only tailored to medication recommendation; its input data was limited to single visits, and the dataset used for pre-training was relatively small, containing 20K patients. 

The latest BERT adaptation, MedBert \citep{Rasmy2021}, had a similar patient representation as BEHRT except that it did not include any temporal information in their model. MedBert excluded age embeddings and visit segment embeddings, along with excluding the  \textbf{SEP} token inserted between visits in favor of including more concept codes. MedBert was trained on EHR data from 20 million patients and introduced a new second learning objective to predict whether the patient had a prolonged length of stay (defined as inpatient visit longer than 7 days). The authors fine-tuned for two disease prediction tasks using a different data source to demonstrate the potential of few-shot learning and showed improved performance. Though a large training data set used and the study showed improvement in the prediction tasks, the lack of temporal information may have limited its potential performance. 

Finally, there are two studies that attempted to incorporate temporal information from structured EHR data \citep{Peng2019, Che2018}. These studies adopted a similar strategy of incorporating time intervals between neighboring clinical events into their models (e.g. two neighboring visits or lab values in time-series data). In this work, we take a  different approach of incorporating time with the introduction of artificial time tokens in CHER-BERT, which will be described in the following section.

\section{Data and Preprocessing} \label{dataset}
\subsection{Data}
EHR data from Columbia University Irving Medical Center-New York Presbyterian Hospital (CUIMC-NYP) was converted into Observational Medical Outcomes Partnership (OMOP), a common data model used to support observational studies and managed by the Observational Health Data Science and Informatics (OHDSI) open-science community \citep{Hripcsak2015}. The CUIMC-NYP OMOP instance includes numerous data and clinical domains, including visits, conditions, procedures, medications, lab tests, vital signs, and problem lists, among others. Data spans from the early 1980s to present day. We used the CUIMC-NYP OMOP to generate training data and downstream prediction cohorts. To pre-train and fine-tune BERTs, we limited the data to three OMOP domains - conditions, procedures, and medications. 

\vspace*{-4.5mm}

\subsection{Data processing and patient representation}
\label{patient_representation}
For each patient, all medical codes were aggregated from three domains and constructed into a sequence chronologically. In order to incorporate temporal information, we inserted an artificial time token (ATT) between two neighboring visits based on their time interval. The following logic was used for creating ATTs based on the following time intervals between visits: 1) if less than 28 days, ATTs take on the form of $W_n$ where n represents the week number ranging from 0-3 (e.g. $W_1$); 2) if between 28 days and 365 days, ATTs are in the form of $M_n$ where n represents the month number ranging from 1-11 e.g $M_{11}$; 3) beyond 365 days then a $LT$ (Long Term) token is inserted. In addition, we added two more special tokens — $VS$ and $VE$ to represent the start and the end of a visit to explicitly define the visit segment, where all the concepts associated with the visit are subsumed by $VS$ and $VE$. Conceptually, a patient can be represented as a list of visits,
\begin{equation*}
\begin{aligned}
P = \{ & VS, \;\; v_1, \;\;  VE, \;\; ATT, \\
        & VS, \;\; v_2, \;\; VE, \;\; ATT, \\
        & VS, \;\; v_3, \;\; VE, \;\; ATT, \\
        &  …, \\
        & VS, \;\; v_i, \;\; VE \}
\end{aligned}
\end{equation*}
where $v_i$ represents the $ith$ visit, and each visit consists of a list of medical concepts $v_i = \{c_{i1}, c_{i2}, c_{i3}, \ldots, c_{ij}\}$. We will refer to EHR patient data representation as patient sequences in the rest of the paper. 

\section{Methods}
\label{sec:methods}
\figureref{fig:bert_architecture_final} shows a high level overview of our adapted BERT architecture.  We used multiple sets of embeddings to represent a patient history including concept embeddings, visit segment embeddings, time embeddings and age embeddings. Concept embeddings were used to capture the numeric representations of the concept codes based on underlying co-occurrence statistics, whereas visit segment embeddings were used to indicate the boundaries of visits (values alternating between A and B). Unlike the previous work, we decided to encode both absolute time (time embeddings) and relative time with respect to visits (age embeddings), due to the finding that certain conditions follow a more seasonal pattern (e.g. flu) while other conditions are more age related (e.g. type 2 diabetes). However, because time and age are numeric values that cannot be directly encoded using standard procedures, we therefore followed the methodology proposed by time2vec \citep{Kazemi2019}. A fourier transform was applied to decompose a sequence of time points into a series of sine functions, which are controlled by learnable parameters in order to adapt to specific training data. We concatenated concept, time and age embeddings together, then fed it through a fully connected (FC) layer to bring it back to the original dimension, which became the temporal concept embeddings input for the BERT architecture. 

\begin{figure*}[htb]
\floatconts
  {fig:bert_architecture_final}
  {\caption{Overview of our BERT architecture on structured EHR data. To distinguish visit boundaries, visit segment embeddings are added to concept embeddings. Next, both visit embeddings and concept embeddings go through a temporal transformation, where concept, age and time embeddings are concatenated together. The concatenated embeddings are then fed into a fully connected layer. This temporal concept embedding becomes the input to BERT. We used the BERT learning objective Masked Language Model as the primary learning objective and introduced an EHR specific secondary learning objective visit type prediction.}}
  {\includegraphics[scale=0.415]{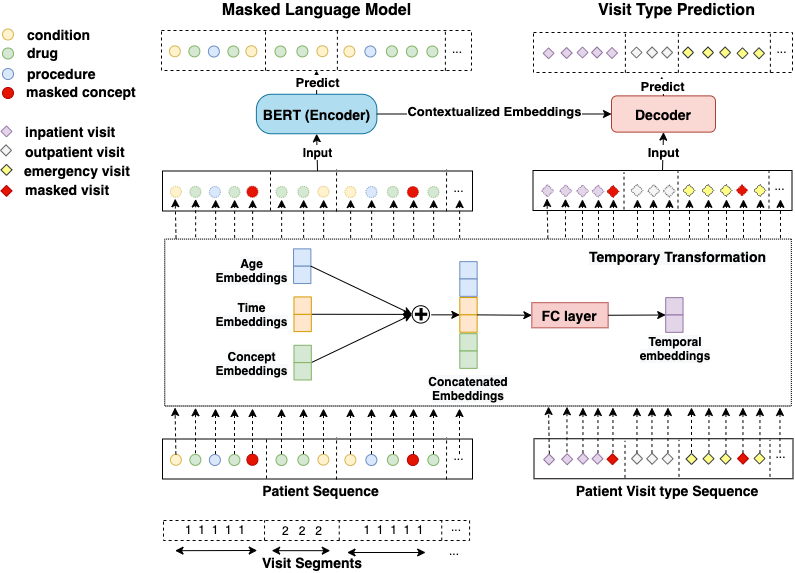}}
  \vspace{-7mm}%Put here to reduce too much white space after your table 
\end{figure*}

For pre-training, we used the core learning objective MLM and followed the standard procedure described in the original BERT paper \citep{Devlin2019}. In addition, we designed a second learning objective named Visit Type Prediction (VTP) to improve BERT’s performance in downstream prediction tasks. VTP was developed based on the observation that different medical concepts are associated with different visit types, and therefore incorporating such domain knowledge may allow BERT to capture additional contextual information. Conceptually, VTP can be thought of as a language translation task, where a sequence of medical concepts are translated to a sequence of visit types (e.g. Inpatient visit v.s. Outpatient visit). To realize this idea, we added a single decoder layer to the BERT architecture to perform VTP. The decoder setup can be summarized as follows: 1) 50\% of tokens in the visit sequence are randomly masked; 2) the visit type sequence undergoes the same temporal transformation as the concept sequence to generate the temporal visit type embeddings; 3) in the decoder, the temporal visit type embeddings and contextualized concept embeddings are combined using multi-headed attention to produce the contextualized visit embeddings; 4) contextualized visit embeddings are used to predict the original visit types for those masked positions in the visit sequence. This second learning objective was trained together with the primary learning objective, MLM. It should be emphasized that the visit sequence is not a list of visits that a patient experienced in the past, but rather, it is a list of visit types constructed from the corresponding concepts in the patient sequence.

\section{Experiments and Results}
\label{sec:results}
\subsection{Experiment setup}
CEHR-BERT was pre-trained using EHR data from a Columbia University Irving Medical Center-New York Presbyterian Hospital (CUIMC-NYP). Patients who had at least one visit and more than 5 data points in their medical history were included, resulting in 2.4M patients and 184.7M clinical data points across OMOP domains - condition, procedure, and medication. The data characteristics can be found in \tableref{tab:summery-stats}. For pre-training BERT, we used 5 encoders and 8 heads with a dropout rate of $0.1$, along with embedding and hidden dimensions of $128$. The context window of 300 tokens rather than the standard 512 was used to construct the patient sequence because 300 is enough to capture more than 90\% of patients' entire medical histories. For those patients who have a sequence of more than 300 codes, we randomly sliced a subset of their sequence (patient history) for pre-training, while patients with less than 300 codes were post-padded with the \textbf{PAD} token. We trained the BERT model for 5 epochs using the Adam optimizer \citep{Kingma2014} with a batch size of 32. The learning rate was initially set to $2e^{-4}$ and the Cosine Annealing LR was used to decay learning rate after every epoch. All training and testing for this study was done on a Linux based server with 768 GB memory, dual Intel(R) Xeon(R) Gold 6138 CPU @ 2.00GHz processors, and two RTX 2080ti GPUs.

\begin{table}[hbt]
\floatconts
  {tab:summery-stats}
  {\caption{Summary statistics of the CUIMC-NYP OMOP instance}}
  {\begin{tabular}{cp{2.5cm}p{2.5cm}}
  \toprule
    & No. of visits per patient & No. of records per patient\\
  \midrule
  mean & 14 & 76\\
  std & 29 & 196\\
  min & 1 & 5\\
  25\% & 2 & 10\\
  50\% & 5 & 24\\
  75\% & 14 & 68\\
  max & 1106 & 31226\\
  \bottomrule
  \end{tabular}}
\end{table}

\subsection{Experiments}
To evaluate the model, we followed previously published standards where we fine-tuned BERT together with a Bi-LSTM layer for a set of disease prediction tasks \citep{Rasmy2021}. Then, we conducted a few-shot learning experiment using the same setup except that a subset of the training data was used for fine-tuning and the model performance was evaluated using the full validation set \citep{McDermott2021}. In these experiments, we included both BEHRT and MedBert as BERT comparators to better understand the relative performance of our model. To understand the contribution of pre-training, we also included a non pre-trained version of CHER-BERT, which will be referred to as R-BERT in the rest of the paper. In addition, several baseline models were evaluated, including logistic regression (LR), XG-boost, and Bi-LSTM with pre-trained time attention concept embeddings \citep{Xiang2019}. Finally, we conducted ablation studies to understand how different temporal components in our adaption could impact the model performance. 

\subsubsection{Disease prediction}
\label{disease_prediction}
Disease prediction is the likelihood of a patient experiencing a condition (disease) in a given time window. \tableref{tab:summery-predictions} shows the 4 prediction tasks including demographics and patient outcomes. Full definitions can be found in section \ref{apd:first}. For feature extraction, a one-year observation window was used prior to the entry of the target cohort by default unless stated otherwise in the prediction definition.

\begin{table*}[hbt]
\floatconts
  {tab:summery-predictions}
  {\caption{Definitions and cohort characteristics of prediction tasks }}
  {\begin{tabular}{lp{3.0cm}p{3.0cm}p{3.0cm}p{3.0cm}}
  \toprule
   & \textit{HF readmission}  & \textit{Discharge home Death} & \textit{T2DM HF} & \textit{Hospitalization}\\
  \midrule
  \abovestrut{2.2ex} Cohort size & $97758$ & $207919$ & $114564$ & $590578$\\
  \abovestrut{2.2ex} Median age & $72$ & $49$ & $61$ & $45$\\
  \abovestrut{2.2ex} Male & $50.30\%$ & $33.23\%$ & $49.66\%$ & $37.46\%$\\
  \abovestrut{2.2ex} Female & $49.71\%$ & $66.77\%$ & $50.34\%$ & $62.54\%$\\
  \abovestrut{2.2ex} Outcome & $24.16\%$ & $4.85\%$ & $9.38\%$& $10.90\%$ \\
  \abovestrut{4.2ex} Description  & 30 days all-cause readmission in HF patients \citep{Golas2018}, see \ref{hf_readmit} for details. & Mortality within 1 year since discharged to home, see \ref{discharged_dead} for details & Life time heart failure prediction since the initial diagnosis of type 2 diabetes mellitus \citep{Rasmy2021}, the medical codes used can be found in \ref{t2dm_hf} & 2 year risk of hospitalization starting from the $3rd$ year since the inital entry into the EHR system \citep{Zhang2018}, see \ref{hospitalization} for details\\
  \bottomrule
  \end{tabular}}
\end{table*}

Extracting features for sequence models (e.g. LSTM and BERTs) was straightforward, where medical concepts were organized in a sequence using chronological order; and artificial tokens and the \textbf{SEP} token were inserted between neighboring visits for CEHR-BERT and BEHRT respectively. The pre-truncate and post-padding strategy was utilized to standardize the size of the inputs for patients who had more/less concepts than the size of the context window. For LR and XG-boost, frequency-based features were constructed by counting the number of occurrences of medical concepts in the observation window. Additionally, medical concepts were rolled up using  ontological hierarchies to reduce dimensionality \citep{Ng2016}, for which a detailed explanation could be found in section \ref{feature_engineering}

For performance evaluation, the 4-fold evaluation was utilized, where in each fold split the data was split into three different sets (75:10:15 split for train/val/test). The sequence models were trained for 10 epochs with early stopping to monitor validation loss with the patience set to 1. On the other hand, since frequency based models (LR and XGB) were not fine-tuned for hyper-parameters, $85\%$ of the data was used for training and the remaining $15\%$ was used for testing directly. The python library sklearn was utilized for training with the default configuration. At the end of each fold, the area under the receiver operating characteristics curve (AUC) was calculated using the test set. In addition, PR-AUC (Precision-Recall) was reported because some of the prediction cohorts had imbalanced outcomes e.g. \textit{Discharge Home Death}.

\tableref{tab:main-results} shows the average AUC and PR-AUC for all models across 4 prediction tasks, where the best value is highlighted in bold. Overall, sequence models outperformed frequency based models, and BERTs performed better than bi-LSTM. Among all models evaluated, CEHR-BERT achieved the best performance in both AUC and PR-AUC across all tasks. In particular, CEHR-BERT is the only model that achieved an AUC of $80$ in \textit{t2dm hf}, and its PR-AUC exceeded the second best performing model MedBert by more than $10$\% (from 0.274 to 0.323). The second and third top performing models were MedBert and BEHRT, whose performances were consistently worse than that of CEHR-BERT. However, these relative performances did not follow the same trend in \textit{hospitalization}, where Bi-LSTM was the second best performing model after CEHR-BERT and outperformed MedBERT and BEHRT. Finally, R-BERT was performing consistently worse than CHER-BERT, suggesting that pre-training played an important role in improving the downstream prediction tasks. 

\begin{table*}[hbt]
\floatconts
  {tab:main-results}
  {\caption{Average AUC and PR-AUC values and standard deviations for three baseline models and three BERT based models across 4 prediction tasks}}
  {\resizebox{\textwidth}{!}{\begin{tabular}{lclllllllll}
  \toprule
     & & LR & XGB & LSTM & R-BERT & BEHRT & MedBert & CEHR-BERT \\
  \midrule
    \multirow{2}{*}{t2dm hf} & PR & 24.8±0.8\% & 24.8±0.5\% & 25.8±0.7\% & 28.1±1.5\% & 27.1±1.7\% & 27.37±0.6\% &  \textbf{32.3±1.0\%}\\
    & AUC & 76.7±0.3\% & 76.5±0.5\% & 77.4±0.5\% & 78.0±0.9\% & 77.5±0.5\% & 78.19±0.1\% & \textbf{80.7±0.6\%}\\
    \multirow{2}{*}{hf readmit} & PR & 36.3±0.7\% & 37.6±1.3\% & 33.3±0.5\% & 37.1±0.6\% & 37.4±0.9\% & 38.0±0.5\% & \textbf{38.6±0.1\%}\\
    & AUC & 64.2±0.7\% & 64.0±0.3\% & 61.7±0.2\% & 65.0±0.5\% & 65.1±0.4\% & 65.8±0.2\% & \textbf{66.3±0.2\%}\\
    \multirow{2}{*}{\makecell{discharge \\home death}} & PR & 46.7±0.7\% & 48.5±0.8\% & 49.1±1.8\% & 46.7±1.2\% & 50.7±0.5\% & 51.4±0.5\% & \textbf{52.7±0.4\%}\\
    & AUC & 93.4±0.1\% & 93.5±0.3\% & 93.8±0.2\% & 93.6±0.2\% & 94.0±0.1\% & 94.2±0.1\% & \textbf{94.6±0.1\%}\\
    \multirow{2}{*}{hospitalization} & PR & 26.9±0.5\% & 29.0±0.3\% & 30.0±0.4\% & 28.0±0.5\% & 29.4±0.2\% & 29.5±0.3\% & \textbf{31.1±0.4\%}\\
    & AUC & 72.9±0.1\% & 74.0±0.1\% & 75.1±0.2\% & 74.4±0.3\% & 74.7±0.1\% & 74.6±0.1\% & \textbf{75.9±0.1\%}\\
  \bottomrule
  \end{tabular}}}
  \vspace{-3mm}%Put here to reduce too much white space after your table 
\end{table*}

\subsubsection{Performance of few-shot learning}
One of the advantages of using pre-trained models is the ability to leverage prior knowledge captured from millions of training examples during the pre-training phase. As a result, BERT could be fine-tuned for downstream tasks using a small training set to achieve decent performance. The setup for few-shot learning was similar to the 4-fold evaluation except that we used a subset of the training data in each fold. Specifically, we predefined a list of training percentages (5\%, 10\%, 20\%, 40\%, and 80\%) with respect to 75\% of the training data in each fold in the experiment, and then iterated through the percentages to conduct a separate 4-fold evaluations, in which a subset of the training data was randomly selected based on the training percentage. Apart from randomly sampling the training data, the other configurations including disease phenotypes, baseline models, the process for train/val/test split, and reporting metrics were identical to the disease prediction tasks. \figureref{fig:transfer_learning} shows that CEHR-BERT is the best performing model in terms of AUC and PR-AUC at different training percentages for \textit{t2dm hf}. Following CEHR-BERT,  MedBert and BEHRT were other top performers compared to LR, XGB and BI-LSTM. CEHR-BERT outperformed the second best model MedBert by the same margin throughout the course of this few-show learning experiment. Noticeably, CEHR-BERT fine-tuned with 5\% of the training data could achieve an AUC of nearly $0.78$ and the PR-AUC of almost $0.29$, whereas other models only achieved the AUCs between 0.60 and 0.76 and PR-AUCs between 0.12 and 0.26 despite using up to 80\% of the training data. Furthermore, the same trend can be observed for other prediction tasks as well, shown in \figureref{fig:discharge_home_death_apd_fig,fig:hospitalization_apd_fig,fig:t2dm_hf_apd_fig}. The only exception is in \textit{hospitalization}, where CEHR-BERT and LSTM were the first and second best performing models; whereas, other BERTs marginally improved the performance compared to frequency based models. 

\subsection{Ablation studies}
To better understand the contributions of time tokens, time/age embeddings and VTP, we conducted a number of ablation studies. We pre-trained several variations of BERT by excluding one component at a time, then conducted the evaluation described in \nameref{disease_prediction}, and reported results for each variation in \tableref{tab:ablation-studies-results}. The best value is highlighted in bold in each row. We discuss the contribution of each component separately in the following sections. We also provided the size of each BERT network and the pre-training time in \tableref{tab:parameter_counts} so the practitioners could choose the appropriate model based on the use-case.

\begin{table*}[hbt]
  \floatconts{tab:ablation-studies-results}
  {\caption{Average AUC and PR-AUC values and standard deviations for ablation studies}}
  {\resizebox{\textwidth}{!}{
  \begin{tabular}{lllllllll}
  \toprule
     & & M-BERT & B-BERT & NS-BERT & NT-BERT & ALT-BERT & V-BERT & CEHR-BERT \\
  \midrule
    \multirow{2}{*}{t2dm hf} & PR & 29.9±1.0\% & 30.6±0.5\% & 31.8±1.3\% & 28.2±0.2\% & 29.3±0.5\% & 28.6±0.8\% & \textbf{32.3±1.0\%}\\
    & AUC & 79.2±0.3\% & 79.5±0.5\% & 80.2±0.6\% & 78.5±0.4\% & 76.7±0.2\% & 78.6±0.1\% & \textbf{80.7±0.6\%}\\
    \multirow{2}{*}{hf readmit} & PR & 34.1±0.6\% & 37.4±0.9\% & 38.3±0.6\% & \textbf{39.3±0.7\%} & 33.3±0.8\% & 38.6±0.4\% & 38.6±0.1\%\\
    & AUC & 62.6±0.3\% & 65.1±0.2\% & 65.8±0.1\% & \textbf{66.4±0.3\%} & 61.6±0.7\% & 65.9±0.2\% & 66.3±0.2\%\\
    \multirow{2}{*}{\makecell{discharge \\home death}} & PR & \textbf{53.1±0.5\%} & 52.0±1.0\% & 52.0±0.7\% & 52.5±0.8\% & 31.6±3.0\% & 52.6±1.3\% & 52.7±0.4\%\\
    & AUC & 94.4±0.3\% & 94.4±0.1\% & 94.3±0.1\% & 94.2±0.3\% & 87.3±0.9\% & 94.4±0.1\% & \textbf{94.6±0.1\%}\\
    \multirow{2}{*}{hospitalization} & PR & 30.0±0.7\% & 30.4±0.5\% & \textbf{31.3±0.7\%} & 30.8±0.6\% & 23.4±0.5\% & 30.6±0.4\% & 31.1±0.4\%\\
    & AUC & 74.9±0.3\% & 75.3±0.3\% & \textbf{76.1±0.2\%} & 75.2±0.2\% & 69.2±0.4\% & 75.3±0.3\% & 75.9±0.1\%\\
  \bottomrule
  \end{tabular}}}
  \footnote{} M-BERT: CEHR-BERT trained on MedBert patient representation\\
  \footnote{} B-BERT: CEHR-BERT trained on BEHRT patient representation\\
  \footnote{} NS-BERT: CEHR-BERT without the second learning objective\\
  \footnote{} NT-BERT: CEHR-BERT without the time/age embeddings\\
  \footnote{} ALT-BERT: modified CEHR-BERT where concept, time and age embeddings are summed together \\
  \footnote{} V-BERT: modified CEHR-BERT where \textit{VS} and \textit{VE} are removed from the patient sequence
\end{table*}

\subsubsection{Assessment of time tokens}
To understand the effectiveness of time tokens, we compared our patient representation described in \nameref{patient_representation} to the existing ones adopted by BEHRT and MedBert. MedBert used a patient representation that only contained the medical concepts and did not include any other artificial tokens. BEHRT used a variation of the M-BERT patient representation, where the \textbf{SEP} token was inserted between neighboring visits. \figureref{fig:patient_representation_comparison} shows an example of these different EHR representations for a patient's medical history. 
To perform a fair comparison, we applied the same architecture (described in \nameref{sec:methods}) to pre-train using these two patient representations, and then we tested the model performances following the same evaluation procedure described in \nameref{disease_prediction}. We will refer to these two BERTs as B-BERT (trained using the BEHRT representation) and M-BERT (trained using the MedBert representation) to distinguish them from others. \tableref{tab:ablation-studies-results} shows that CEHR-BERT improved the performances in all tasks as compared to the other patient representations except that M-BERT performed slightly better in terms of PR-AUC in \textit{discharge home death}. Our results indicate that embedding ATTs can effectively capture the temporal information of structured EHR data. One plausible explanation is that ATTs may be treated like any other tokens such that when BERT utilizes them in the self-attention mechanism, those tokens function like a place holder for preserving temporal information, which is then propagates through to the last encoder. In addition, we trained another CEHR-BERT variation where we removed \textit{VS} and \textit{VE} tokens and only kept the ATT tokens, (which we will be refer to as V-BERT). The performance of V-BERT dropped compared to CEHR-BERT, and was comparable to that of B-BERT (\textit{SEP} token inserted between visits), suggesting that \textit{VS} and \textit{VE} play an important role in explicitly defining the boundary of a visit, so that ATT tokens would just function like a regular \textit{SEP} token without their presence. 

\subsubsection{Assessment of visit type prediction}
VTP was designed in this study as a substitute for the original second learning objective NSP to leverage the unique characteristics of structured EHR data. 
To understand its contribution, we excluded VTP and pre-trained BERT, and then followed the standard evaluation. This BERT will be referred to as NS-BERT. The comparison between NS-BERT and CEHR-BERT in \tableref{tab:ablation-studies-results} demonstrates the improved performances across all prediction tasks attributed to VTP with the exception of \textit{hospitalization}, where NS-BERT slightly outperformed CEHR-BERT but both remained largely similar. Such minor differences may be due to random variation of 4-fold evaluation. The underlying assumption of VTP was that concepts associated with different visit types follow different distributions. Therefore, incorporating VTP into BERT could help learn the representation of concepts. To better understand this, we reported the top 10 most frequent concepts stratified by visit type and domain in \tableref{tab:top_10_most_frequent_concepts_part1} in \nameref{apd:second}, which shows that the concept distributions are indeed distinct between visit types. For example, the most frequent concepts associated with inpatient visits relate to childbirth or severe conditions that require immediate hospitalization, while the top outpatient concepts normally pertain to chronic conditions such as hypertension and type 2 diabetes that may not require immediate medical attention but long-term management. Currently, the threshold for masking visit tokens is set to 50\%, but we plan to investigate different thresholds to optimize the performance of BERT. 

\subsubsection{Assessment of time/age embeddings}
At the input layer, concept, time, and age embeddings are concatenated together, and fed into a FC layer to form temporal concept embeddings, which is then used as the input for CEHR-BERT. To understand the impact of this transformation, we pre-trained a BERT named ALT-BERT, where we summed time embeddings, age embeddings and concept embeddings to form temporal concept embeddings instead of using a FC layer. \tableref{tab:ablation-studies-results} shows that the performance of ALT-BERT was lower compared to that of CEHR-BERT, suggesting that our temporal concept embeddings transformation is more effective than the summation of time/age/concept embeddings. This confirms existing empirical evidence that summing is more rigid than feeding a concatenated product into another the FC layer in terms of fitting the model to the training data. 

Furthermore, we wanted to know the impact of using time/age embeddings. Therefore we pre-trained another BERT named NT-BERT, where we disabled all components related to time/age embeddings and only used the concept embeddings. We added the positional encoder (used by \citet{Rasmy2021, Li2020}) to give the model a basic sense of temporality. \tableref{tab:ablation-studies-results} shows that
without time/age embeddings, NT-BERT did perform slightly better in \textit{hf readmit} than CEHR-BERT; however, CEHR-BERT outperformed NT-BERT in all other tasks. In particular, PR-AUC for NT-BERT in \textit{t2dm hf} dropped by 4\% compared to CEHR-BERT (from 32.3±1.0\% to 28.2±0.2\%). Therefore, the benefit of performing the transformation for generating temporal concept embeddings likely outweighs the cost. 

\section{Discussion}
\subsection{Understanding artificial time tokens}
To understand the functional role of ATTs, we extracted the base embeddings of ATTs and computed 2D features using PCA for visualization. \figureref{fig:time_token_visualization} shows that ATTs were arranged from right to left in increasing order of time intervals, where the rightmost and leftmost tokens represent the shortest and longest time intervals respectively. Specifically, the ATT week tokens (including $W_0$, $W_1$, and $W_2$) formed a cluster on the right bottom, and month tokens (in the form of $M_n$ e.g. $M_1$) seemed to present a linear relationship. In addition, the VS and VE tokens were located far away from time tokens because they represent the start and the end of a visit rather than time intervals, corroborating that their semantics are fundamentally different from time tokens. The analysis supported our assumption that BERT can derive semantics of ATTs and show meaningful relationships between them, thereby learning the hidden knowledge that different time tokens are associated with different groups of concepts.

\subsection{Inflated patient sequence}
Due to the use of ATTs, we artificially inflated the length of patient sequences, and as a result involuntarily cut-off some records from patient sequences. Whereas, in other sequence models (M-BERT, and Bi-LSTM) we used patient sequences as-is. We calculated the median length (defined as the median number of concepts) as well as 95\% sequence length (defined as the number of concepts at 95\% percentile) of the patient sequences for different BERT representations across 4 disease prediction tasks, which is shown in \tableref{tab:inflated-patient-sequences}. Among all tasks, patient sequences for \textit{t2dm hf} and \textit{hospitalization} patients have the shortest patient sequences; whereas, \textit{hf readmission} has the richest medical history. Among all patient representations, CEHR-BERT has the longest sequence due to the use of ATTs, and B-BERT has the second longest sequence due to the use of \textbf{SEP}. This raised the question of whether cutting off early records in patient sequences could have contributed to the performance boost in evaluations. To address this concern, we re-ran the \textit{hf readmission} analysis using a shorter observation window of $180$ days for all models except CEHR-BERT. The motivation was to simulate a scenario where early records were omitted and only the latest records were included. The performance of this modified evaluation was reported in \tableref{tab:hf_readmission_180_days_results}. It showed that model performances did not improve but dropped when compared to the original evaluation in \tableref{tab:main-results} and cutting off the early records in patient sequences did not improve the performance. This comparison suggests the performance gain by CEHR-BERT is attributed to the use of \textbf{ATT} tokens. 

\section{Future work}
In this work, we used a context window of 300 for pre-training and conducting experiments. Although we can increase the context window to the BERT sequence limit of 512, patients may have more clinical concepts in their medical histories than the sequence length supported by BERT. To address this limitation, we plan to employ a sliding window strategy, in which we can apply BERT to extract the contextualized embedding representations for each region scanned by the sliding window. Then the regional representations could be combined via a 1D convolution layer or LSTM to generate the final patient representation. Although VTP seems to have improved the results by a small margin, it is not clear whether the improvement could be attributed to random variation, so we need to investigate VTP in the follow-up analysis. Furthermore, we want to include labs to the patient sequence as labs offer an extensive amount of useful information. However, embedding the lab data would require a different strategy as labs are continuous features unlike the discrete data types included in this study. Finally, we will investigate whether or not our methods of incorporating time could be generalized for other models as well. 

\section{Conclusion}
To the best of our knowledge, this is the first study that focuses on incorporating time into BERT for use on structured EHR data, including multiple elements for representing temporal information and leveraging multiple domains of clinical care. CEHR-BERT outperforms existing state of the art BERT-based approaches across a number of different prediction tasks. Based on our results, incorporating time tokens into the patient sequence and combining time/age embeddings with concept embeddings seem to synergistically boost the performance more than any individual modification alone. Therefore, including both into the final CEHR-BERT architecture seems to be the most effective way of capturing the temporal information of structured EHR data. This study was developed using the OMOP common data model, and as a result can be expanded to run on the OHDSI network in order to replicate these results beyond a single database.

\acks{This work was sponsored by the National Center for Advancing Translational Sciences grant 1U01TR002062 and the Director’s Office of National Institutes of Health grants 5U2COD023196 and 3UG3OD023183.}

\clearpage

\bibliography{att-bert}

\clearpage

\appendix
\section{Prediction Tasks}\label{apd:first}
A prediction task can be phrased as the following, “among a particular group of people, who will go on and experience some event”. One can think of this problem as defining a target cohort that represents the initial group of people, and an outcome cohort that represents the subset of the initial group who will experience a particular event, e.g. among the type 2 diabetes patients, who will go on and develop heart failure. Both target and outcome cohorts can be defined as a group of people who satisfy certain inclusion criteria for a certain period of time. Typically, a cohort definition includes a cohort entry event and a set of inclusion criteria (an exclusion criterion can be thought of as an inclusion criterion with 0 occurrence). Specifically, the cohort entry event defines the index date, at which the patients enter the cohort, and the inclusion criteria add more constraints to the cohort if applicable, such as the requirements of certain diagnosis, medications, procedures or temporal relationships among criteria, and etc. In addition, a prediction window needs to be specified for generating the ground truth labels for the given target and outcome cohorts, if the outcome index date falls between the index date of the target cohort and the prediction window, we will declare the case to be positive, and otherwise negative.

In terms of prediction time range, we use observation window, hold-off window and prediction window to collect data in different time periods. One patient's medical history is built on an event sequence. Each event represents one medical engagement within a visit. It could be condition occurrence, medication exposure, procedure occurrence and measurement etc. In this paper, we only focus on the first three types of events. We extract features using data from the observation window followed by a hold-off window to avoid same target concepts being included into feature construction. The prediction window is right after the index event. And the goal is to predict the occurrence of any outcome event. With the datasets, we could train, test and validate the model. \figureref{fig:cohort_definition_prediction_window} visualizes the cohort definition that is constructed based on all events in patient medical history.

\begin{figure}[htbp]
\floatconts
  {fig:cohort_definition_prediction_window}
  {\caption{Cohort Definition and Prediction Windows}}
  {\includegraphics[width=1.0\linewidth]{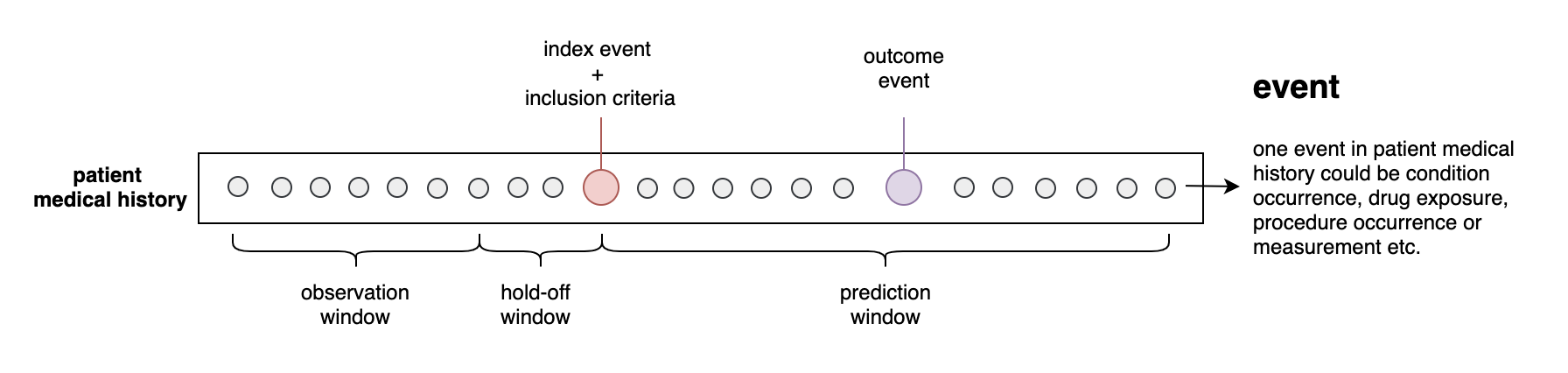}}
\end{figure}

We validate our model by applying it to 4 downstream prediction problems.  \tableref{tab:cohort_definition_parameters} lists the cohort definition parameters for the prediction tasks.

\begin{table*}[hbtp]
\floatconts
  {tab:cohort_definition_parameters}
  {\caption{Cohort Definition Parameters}}
  {\footnotesize
  \begin{tabular} {p{2cm}|p{2cm}|p{2.5cm}|p{1.5cm}|p{1.5cm}|p{1.5cm}|p{1.5cm}}
  \toprule
    cohort name & index event & inclusion criteria & outcome event & observation window & hold-off \newline window & prediction window\\
  \midrule
    t2dm hf & Type 2 Diabetes Mellitus condition occurrence or medication exposure & No pre-existing diabetes, type 1 diabetes, diabetes insipidus, gestational diabetes, secondary diabetes, neonatal diabetes & Heart Failure & unbounded & 0 & unbounded \\
  \hline
    hf readmit & Heart Failure condition occurrence & At least one HF treatment, lab test or medication exposure & Readmission \newline (In-patient visit) & 360 days & 0 & 30 days \\
  \hline
    discharge home death & Discharge to home & Following an in-patient visit & Death & 360 days & 0 & 360 days \\
  \hline
    hospitalization & EHR start & Number of visits between 2 and 30 & In-patient visit & 540 days & 180 days & 720 days \\
    
  \bottomrule
  \end{tabular}}
\end{table*}

\pagebreak
\subsection{Feature Engineering} 
\label{feature_engineering}
Frequency based features were used for LR and XG-boost,  and generated as the following 1) International Classification of Diseases (ICD) ICD-9 or ICD-10 codes were rolled up to 2 or 3 digit codes for procedure and diagnosis depending on where the dot character is in the code e.g. 44.98 and I50.0; 2) for procedure records encoded by Current Procedural Terminology (CPT), they were rolled up to the second level from the root; 3) all medications were rolled up to the ingredient level; 4) for the codes that couldn't be rolled up, the original codes were kept; 5) the frequencies of the rolled-up codes were calculapted within the observation window. 

\subsection{Type 2 Diabetes Mellitus patients who developed Heart Failure}
\label{t2dm_hf}
The target cohort consists of patients who had T2DM (type 2 diabetes mellitus) in the medical history. The index event of T2DM is patient encounters with condition concept ids or medication exposures of anti-diabetes medications. We also exclude any patients with pre-existing T2DM, type 1 diabetes, diabetes insipidus, gestational diabetes, secondary diabetes and neonatal diabetes. 

The outcome cohort is a subset of the target cohort who developed heart failure during the prediction window. We defined Heart Failure cohort as patients who were diagnosed with heart failure, at least one lab test with high BNP results, received any treatment including mechanical circulatory support, artificial heart associated procedure, diuretic agent, vasoactive agent or dialysis procedure.

For each of the criteria, we construct a concept set with a group of OMOP concept ids. In \tableref{tab:t2dm_cohort_related_concepts} \& \tableref{tab:heart_failure_related_concepts}, we list all related concept sets and concept ids from OMOP vocabulary.

\begin{table*}[hbtp]
\floatconts
  {tab:t2dm_cohort_related_concepts}
  {\caption{T2DM Cohort Related Concepts}}
  {\begin{tabular} {p{3cm}|p{4cm}|p{7cm}}
  \toprule
     \multicolumn{1}{l}{\bf Domain}
      & \multicolumn{1}{l}{\bf Concept Set} 
         & \multicolumn{1}{l}{\bf OMOP concept ids} \\
  \midrule
    \multirow{7}{*}{Condition} & \multicolumn{1}{L{6cm}}{T2DM} & \multicolumn{1}{L{7cm}}{443238, 201820, 442793, 4016045} \\\cline{2-3}
                               & \multicolumn{1}{L{6cm}}{pre-existing T2DM} & \multicolumn{1}{L{7cm}}{40769338,43021173,42539022, 46270562} \\\cline{2-3}
                               & \multicolumn{1}{L{6cm}}{type 1 diabetes} & \multicolumn{1}{L{7cm}}{201254, 4019513, 40484648} \\\cline{2-3}
                               & \multicolumn{1}{L{6cm}}{diabetes insipidus} & \multicolumn{1}{L{7cm}}{30968, 438476} \\\cline{2-3}
                               & \multicolumn{1}{L{6cm}}{gestational diabetes} & \multicolumn{1}{L{7cm}}{4058243} \\\cline{2-3}
                               & \multicolumn{1}{L{6cm}}{secondary diabetes} & \multicolumn{1}{L{7cm}}{195771} \\\cline{2-3}
                               & \multicolumn{1}{L{6cm}}{neonatal diabetes} & \multicolumn{1}{L{7cm}}{193323} \\\cline{2-3}
  \hline
    \multirow{19}{*}{Medication} & \multicolumn{1}{L{6cm}}{Metformin} & \multicolumn{1}{L{7cm}}{1503297} \\\cline{2-3}
                           & \multicolumn{1}{L{6cm}}{Chlorpropamide} & \multicolumn{1}{L{7cm}}{1594973} \\\cline{2-3}
                           & \multicolumn{1}{L{6cm}}{Glimepiride} & \multicolumn{1}{L{7cm}}{1597756} \\\cline{2-3}
                           & \multicolumn{1}{L{6cm}}{Glyburide} & \multicolumn{1}{L{7cm}}{1559684} \\\cline{2-3}
                           & \multicolumn{1}{L{6cm}}{Glipizide} & \multicolumn{1}{L{7cm}}{1560171} \\\cline{2-3}
                           & \multicolumn{1}{L{6cm}}{Tolbutamide} & \multicolumn{1}{L{7cm}}{1502855} \\\cline{2-3}
                           & \multicolumn{1}{L{6cm}}{Tolazamide} & \multicolumn{1}{L{7cm}}{1502809} \\\cline{2-3}
                           & \multicolumn{1}{L{6cm}}{Pioglitazone} & \multicolumn{1}{L{7cm}}{1525215} \\\cline{2-3}
                           & \multicolumn{1}{L{6cm}}{Rosiglitazone} & \multicolumn{1}{L{7cm}}{1547504} \\\cline{2-3}
                           & \multicolumn{1}{L{6cm}}{Sitagliptin} & \multicolumn{1}{L{7cm}}{1580747} \\\cline{2-3}
                           & \multicolumn{1}{L{6cm}}{Saxagliptin} & \multicolumn{1}{L{7cm}}{40166035} \\\cline{2-3}
                           & \multicolumn{1}{L{6cm}}{Alogliptin} & \multicolumn{1}{L{7cm}}{43013884} \\\cline{2-3}
                           & \multicolumn{1}{L{6cm}}{Linagliptin} & \multicolumn{1}{L{7cm}}{40239216} \\\cline{2-3}
                           & \multicolumn{1}{L{6cm}}{Repaglinide} & \multicolumn{1}{L{7cm}}{1516766} \\\cline{2-3}
                           & \multicolumn{1}{L{6cm}}{Nateglinide} & \multicolumn{1}{L{7cm}}{1502826} \\\cline{2-3}
                           & \multicolumn{1}{L{6cm}}{Miglitol} & \multicolumn{1}{L{7cm}}{1510202} \\\cline{2-3}
                           & \multicolumn{1}{L{6cm}}{Linagliptin} & \multicolumn{1}{L{7cm}}{40239216} \\\cline{2-3}
                           & \multicolumn{1}{L{6cm}}{Acarbose} & \multicolumn{1}{L{7cm}}{1529331} \\\cline{2-3}
                           & \multicolumn{1}{L{6cm}}{Insulin} & \multicolumn{1}{L{7cm}}{35605670, 35602717, 1516976, 1502905, 46221581, 1550023, 35198096, 42899447, 1544838, 1567198, 35884381, 1531601, 1588986, 1513876, 19013951, 1590165, 1596977, 1586346, 19090204, 1513843, 1513849, 1562586, 19090226, 19090221, 1586369, 19090244, 19090229, 19090247, 19090249, 19090180, 19013926, 19091621, 19090187} \\
    
  \bottomrule
  \end{tabular}}
\end{table*}

\begin{table*}[hbtp]
\floatconts
  {tab:heart_failure_related_concepts}
  {\caption{Heart Failure Cohort Related Concepts}}
  {\begin{tabular} {p{3cm}|p{4cm}|p{7cm}}
  \toprule
     \multicolumn{1}{l}{\bf Domain}
      & \multicolumn{1}{l}{\bf Concept Set} 
         & \multicolumn{1}{l}{\bf OMOP concept ids} \\
  \midrule
    \multirow{1}{*}{Condition} 
        & \multicolumn{1}{L{6cm}}{Heart Failure} & \multicolumn{1}{L{7cm}}{316139} \\\cline{2-3}
  \hline
    \multirow{2}{*}{Medication} 
        & \multicolumn{1}{L{6cm}}{Diuretic Agent} & \multicolumn{1}{L{7cm}}{4186999, 956874, 942350, 987406, 932745, 1309799, 970250, 992590, 907013} \\\cline{2-3}
        & \multicolumn{1}{L{6cm}}{Vasoactive Agent} & \multicolumn{1}{L{7cm}}{1942960} \\\cline{2-3}
  \hline
    \multirow{2}{*}{Measurement} 
        & \multicolumn{1}{L{6cm}}{High B-type Natriuretic Peptide (BNP) $>$ 500 pg/mL} & \multicolumn{1}{L{7cm}}{4307029} \\\cline{2-3}
        & \multicolumn{1}{L{6cm}}{NT-proBNP $>$ 2000 pg/mL} & \multicolumn{1}{L{7cm}}{1594973} \\\cline{2-3}
  \hline
    \multirow{1}{*}{Procedure}
        & \multicolumn{1}{L{6cm}}{Mechanical Circulatory Support} & \multicolumn{1}{L{7cm}}{45888564, 4052536, 4337306, 2107514, 45889695, 2107500, 45887675, 43527920, 2107501, 45890116, 40756954, 4338594, 43527923, 40757060, 2100812} \\\cline{2-3}
        & \multicolumn{1}{L{6cm}}{Artificial Heart Associated Procedure} &
        \multicolumn{1}{L{7cm}}{4144390, 4150347, 4281764, 725038, 725037, 2100816, 2100822, 725039, 2100828, 4337306, 4140024, 4146121, 4060257, 4309033, 4222272, 4243758, 4241906, 4080968, 4224193, 4052537, 4050864} \\
  \bottomrule
  \end{tabular}}
\end{table*}

\subsection{Heart Failure patients who were readmitted within 30 days} 
\label{hf_readmit}
The target cohort contains patients who were admitted into hospital due to heart failure . The index event is an inpatient visit with a heart failure diagnosis (316139). Patients in the target cohort who were readmitted into hospital within 30 days will be in the outcome cohort. The concept\_ids of inpatient visits are 9201 and 262. In this case, the prediction window is 30 days.

\subsection{Patients who were discharged and died within one year} 
\label{discharged_dead}
The target cohort is patients who had an inpatient visit and were discharged to home. The outcome cohort is patients who died within one year after being discharged. The index event is inpatient visit with visit concept\_id as 9201 or 262 and discharge\_to\_concept\_id is home or other nursing facilities. The outcome event is death. The prediction window is 360 days.

\subsection{Hospitalization} 
\label{hospitalization}
The hospitalization cohort has a special structure showed in \figureref{fig:hospitalization_cohort_definition_prediction_window} than the generalized cohort structure. The index event is when the patient had the first visit in the hospital. And the observation window is post the index event. The outcome event is an inpatient visit. We only include patients who had visit occurrences between 2 to 30 during the observation window to make sure patients had enough data points and also remove any outliers.
\begin{figure}[htbp]
\floatconts
  {fig:hospitalization_cohort_definition_prediction_window}
  {\caption{Hospitalization Cohort Definition and Prediction Windows}}
  {\includegraphics[width=1.0\linewidth]{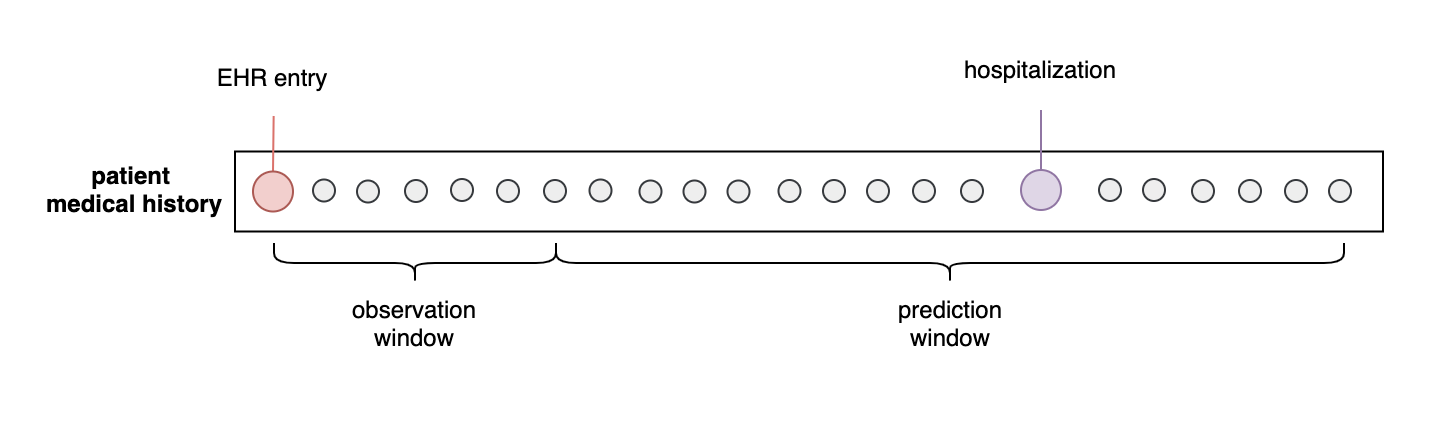}}
\end{figure} 

\section{Additional Figures and Analyses}\label{apd:second}

\begin{figure*}[htb]
\floatconts
  {fig:patient_representation_comparison}
  {\caption{EHR Representation of patient medical history}}
  {\includegraphics[scale=0.53]{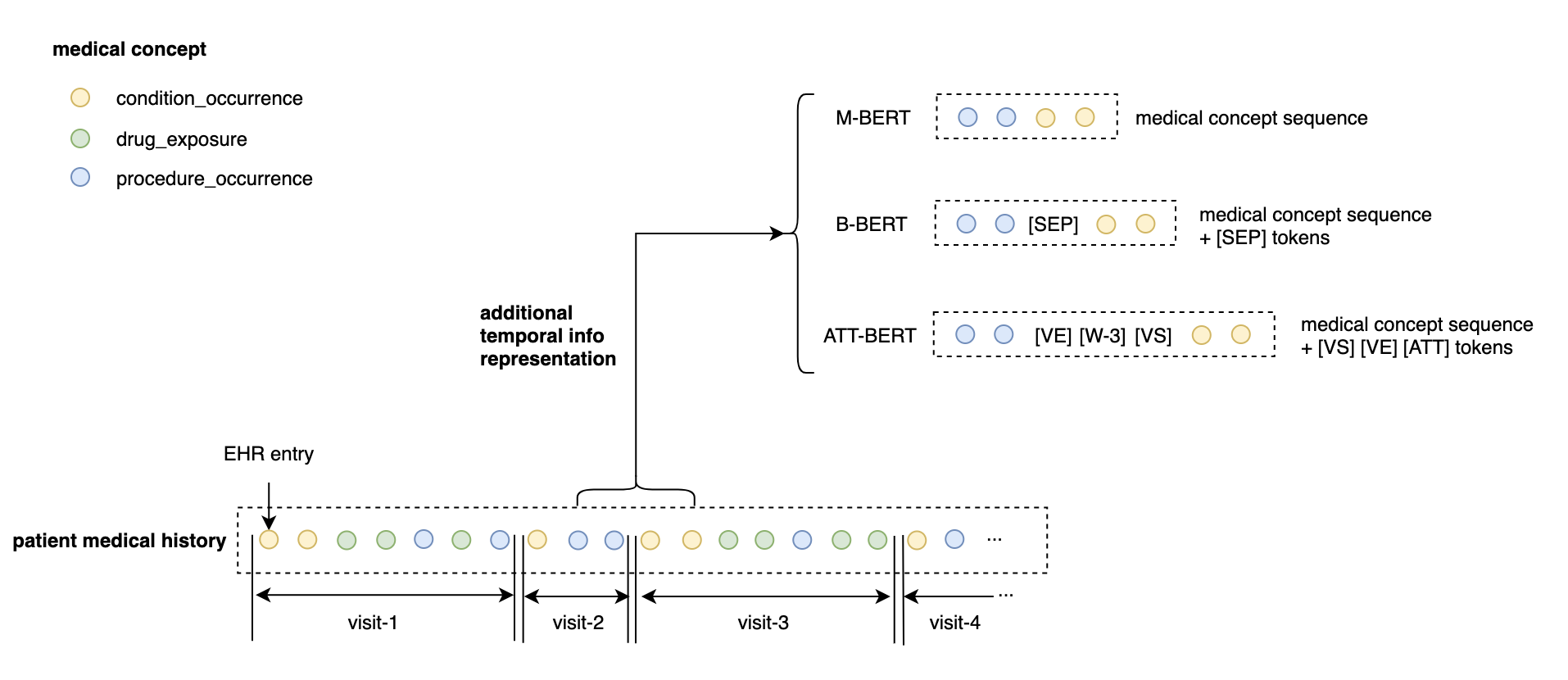}}
\end{figure*}

\begin{figure*}[hbt]
\floatconts
  {fig:time_token_visualization}
  {\caption{2d visualization of the Artificial Time Tokens added to CEHR-BERT. The base embeddings of those tokens were extracted from CEHR-BERT, and PCA was run to extract the 2d features for visualization. }}
  {\includegraphics[scale=0.6]{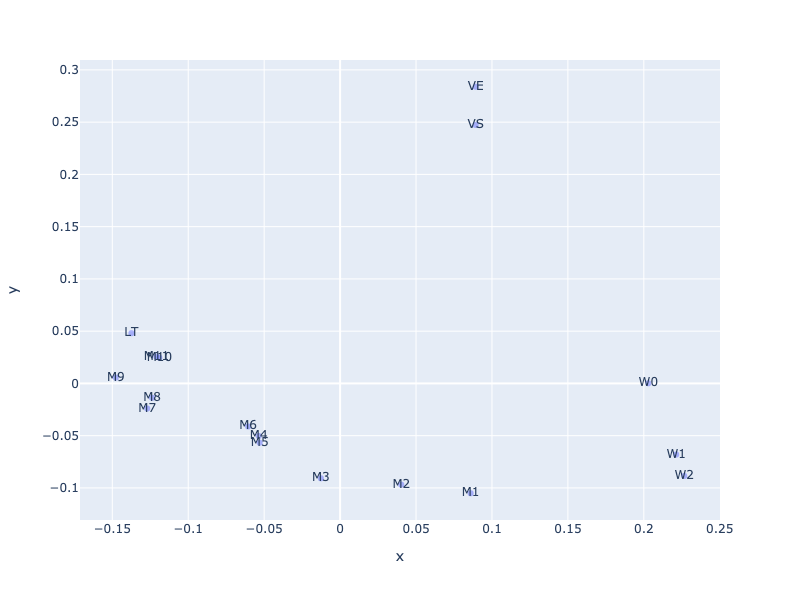}}
\end{figure*}

\begin{figure*}[hbt]
\floatconts
  {fig:transfer_learning}
  {\caption{AUC and PR-AUC at different training percentages for all the models for few-shot learning task for \textit{hf readmission} are plotted against}}
  {\includegraphics[scale=0.5]{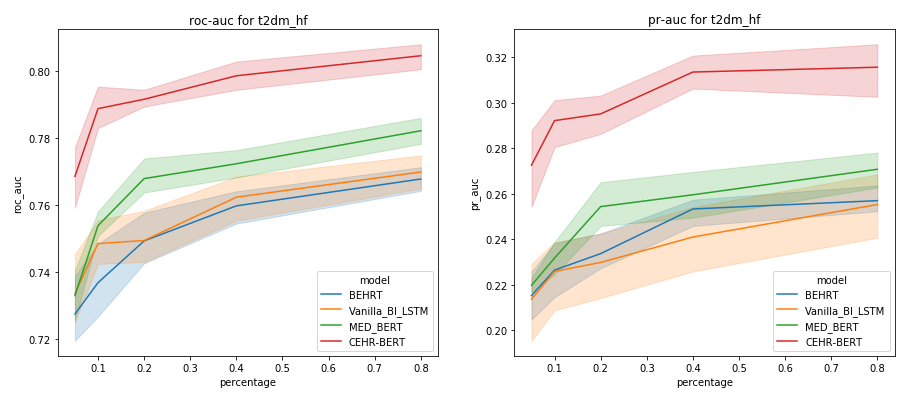}} 
\end{figure*}

\begin{figure*}[hbt]
\floatconts
  {fig:discharge_home_death_apd_fig}
  {\caption{AUC and PR-AUC at different training percentages for all the models for few-shot learning task for \textit{discharge home death} are plotted against}}
  {\includegraphics[scale=0.5]{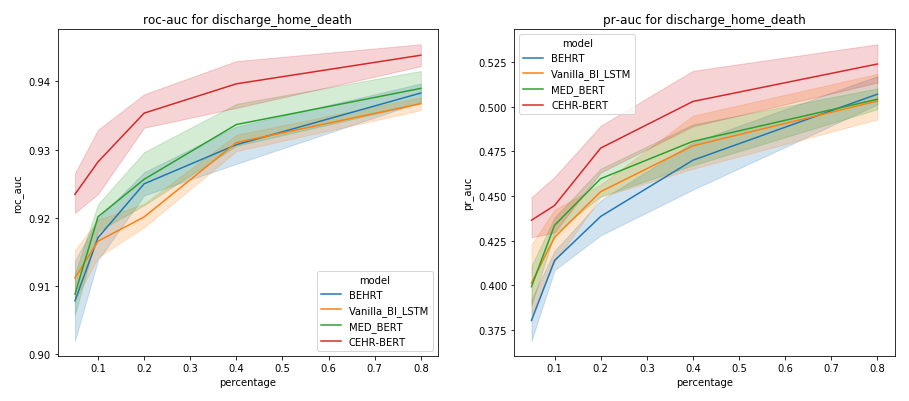}}
\end{figure*}

\begin{figure*}[hbt]
\floatconts
  {fig:hospitalization_apd_fig}
  {\caption{AUC and PR-AUC at different training percentages for all the models for few-shot learning task for \textit{hospitalization} are plotted against}}
  {\includegraphics[scale=0.5]{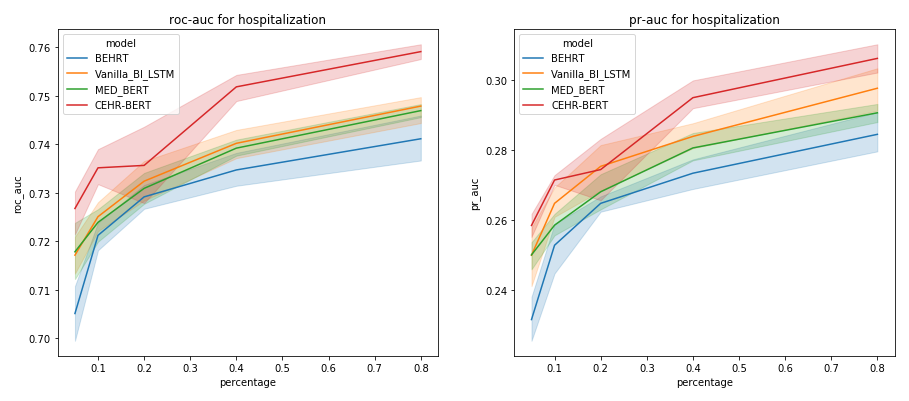}}
\end{figure*}

\begin{figure*}[hbt]
\floatconts
  {fig:t2dm_hf_apd_fig}
  {\caption{AUC and PR-AUC at different training percentages for all the models for few-shot learning task for \textit{hf readmission} are plotted against}}
  {\includegraphics[scale=0.5]{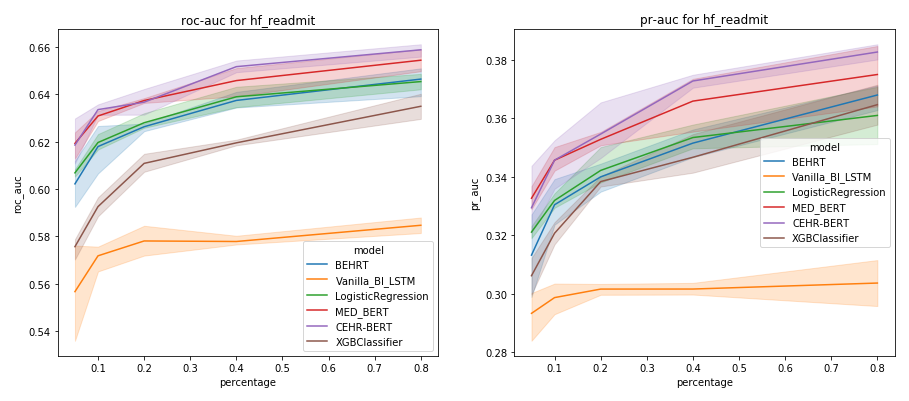}}
\end{figure*}

\begin{table*}[htp]
\floatconts
  {tab:hf_readmission_180_days_results}
  {\caption{Average AUC and PR-AUC values for LR, XGB, LSTM, BEHRT, and MedBert for \textit{hf readmission} using the 180-day observation window}}
  {\begin{tabular}{llllllll}
  \toprule
     & LR & XGB & LSTM & BEHRT & MedBert \\
  \midrule
    PR-AUC & 36.6±0.7\% & 37.1±0.6\% & 36.3±0.1\% & 37.2±0.3\% & 31.7±0.4\%\\
    AUC & 64.7±0.3\% & 64.1±0.7\% & 64.2±0.4\% & 65.0±0.3\% & 59.5±0.1\%\\
  \bottomrule
  \end{tabular}}
\end{table*}

\begin{table*}[htbp]
\floatconts
  {tab:parameter_counts}
  {\caption{Trainable and Non trainable parameters counts for M-BERT, B-BERT, NS-BERT, ALT-BERT, V-BERT and CEHR-BERT.}}
  {\begin{tabular}{llllllll}
  \toprule
     & M-BERT & B-BERT & NS-BERT & ALT-BERT & V-BERT & CEHR-BERT \\
  \midrule
    Trainable Parameters & 9083060 & 9083060 & 8811380 & 9044596 & 9085108 & 9085364\\
    NonTrainable Parameters & 0 & 0 & 0 & 0 & 0 & 0\\
    Time to pre-train per epoch & 8.5h & 8.5h & 7.5h & 8.5h & 8.5h & 8.5h &\\
  \bottomrule
  \end{tabular}}
\end{table*}

\begin{table*}[hbt]
\floatconts
  {tab:inflated-patient-sequences}
  {\caption{Patient sequence lengths for 3 BERT representations across 4 disease predictions}}
  {\begin{tabular}{llllll}
  \toprule
   & & \textit{HF readmission}  & \textit{Discharge home Death} & \textit{T2DM HF} & \textit{Hospitalization}\\
  \midrule
  \multirow{3}{*}{\makecell{median length}} & CEHR-BERT & 123 & 64 & 21 & 19 \\
  & b-bert & 98 & 47 & 15 & 12 \\
  & m-bert & 86 & 38 & 13 & 9 \\
  \multirow{3}{*}{\makecell{95\% length}} & CEHR-BERT   & 608 & 330 & 366 & 108 \\
  & b-bert  & 520 & 257 & 246 & 78 \\
  & m-bert  & 481 & 223 & 189 & 66 \\
  \bottomrule
  \end{tabular}}
\end{table*}

\begin{table*}[hbtp]
\floatconts
  {tab:top_10_most_frequent_concepts_part1}
  {\caption{Top 10 most frequent condition concepts associated with each visit type}}
  {\begin{tabular}{p{3cm}|p{0.8cm}|p{6cm}|p{2cm}|p{2cm}}
  \toprule
     \multicolumn{1}{p{3cm}}{Visit Type}
      & \multicolumn{1}{p{0.8cm}}{Rank} 
      & \multicolumn{1}{p{6cm}}{Condition Concepts}
      & \multicolumn{1}{p{2cm}}{Patient Count}
      & \multicolumn{1}{p{2cm}}{Percentage of the Visit Type Patients}\\
  \midrule
    \multirow{2}{3cm}{Outpatient Visit} 
        & \multicolumn{1}{L{0.8cm}}{1} & \multicolumn{1}{L{6cm}}{Essential hypertension} & \multicolumn{1}{L{2cm}}{310794} & \multicolumn{1}{L{2cm}}{13.00\%} \\\cline{2-5}
        & \multicolumn{1}{L{0.8cm}}{2} & \multicolumn{1}{L{6cm}}{Abdominal pain} & \multicolumn{1}{L{2cm}}{172393} & \multicolumn{1}{L{2cm}}{7.21\%} \\\cline{2-5}
        & \multicolumn{1}{L{0.8cm}}{3} & \multicolumn{1}{L{6cm}}{Chest pain} & \multicolumn{1}{L{2cm}}{164499} & \multicolumn{1}{L{2cm}}{6.88\%} \\\cline{2-5}
        & \multicolumn{1}{L{0.8cm}}{4} & \multicolumn{1}{L{6cm}}{Hyperlipidemia} & \multicolumn{1}{L{2cm}}{160068} & \multicolumn{1}{L{2cm}}{6.70\%} \\\cline{2-5}
        & \multicolumn{1}{L{0.8cm}}{5} & \multicolumn{1}{L{6cm}}{Finding related to pregnancy} & \multicolumn{1}{L{2cm}}{140044} & \multicolumn{1}{L{2cm}}{5.86\%} \\\cline{2-5}
        & \multicolumn{1}{L{0.8cm}}{6} & \multicolumn{1}{L{6cm}}{Joint pain} & \multicolumn{1}{L{2cm}}{117134} & \multicolumn{1}{L{2cm}}{4.90\%} \\\cline{2-5}
        & \multicolumn{1}{L{0.8cm}}{7} & \multicolumn{1}{L{6cm}}{Dyspnea} & \multicolumn{1}{L{2cm}}{114580} & \multicolumn{1}{L{2cm}}{{4.79\%}} \\\cline{2-5}
        & \multicolumn{1}{L{0.8cm}}{8} & \multicolumn{1}{L{6cm}}{Low back pain} & \multicolumn{1}{L{2cm}}{114031} & \multicolumn{1}{L{2cm}}{{4.77\%}} \\\cline{2-5}
        & \multicolumn{1}{L{0.8cm}}{9} & \multicolumn{1}{L{6cm}}{Pure hypercholesterolemia} & \multicolumn{1}{L{2cm}}{113437} & \multicolumn{1}{L{2cm}}{{4.75\%}} \\\cline{2-5}
        & \multicolumn{1}{L{0.8cm}}{10} & \multicolumn{1}{L{6cm}}{Unplanned pregnancy} & \multicolumn{1}{L{2cm}}{110584} & \multicolumn{1}{L{2cm}}{{4.63\%}} \\\cline{2-5}
  \hline
    \multirow{2}{3cm}{Inpatient Visit} 
        & \multicolumn{1}{L{0.8cm}}{1} & \multicolumn{1}{L{6cm}}{Single live birth} & \multicolumn{1}{L{2cm}}{279820} & \multicolumn{1}{L{2cm}}{35.01\%} \\\cline{2-5}
        & \multicolumn{1}{L{0.8cm}}{2} & \multicolumn{1}{L{6cm}}{Essential hypertension} & \multicolumn{1}{L{2cm}}{150002} & \multicolumn{1}{L{2cm}}{{18.77\%}} \\\cline{2-5}
        & \multicolumn{1}{L{0.8cm}}{3} & \multicolumn{1}{L{6cm}}{Finding related to pregnancy} & \multicolumn{1}{L{2cm}}{121545} & \multicolumn{1}{L{2cm}}{{15.21\%}} \\\cline{2-5}
        & \multicolumn{1}{L{0.8cm}}{4} & \multicolumn{1}{L{6cm}}{Postpartum finding} & \multicolumn{1}{L{2cm}}{100345} & \multicolumn{1}{L{2cm}}{{12.55\%}} \\\cline{2-5}
        & \multicolumn{1}{L{0.8cm}}{5} & \multicolumn{1}{L{6cm}}{Delivery normal} & \multicolumn{1}{L{2cm}}{66666} & \multicolumn{1}{L{2cm}}{{8.34\%}} \\\cline{2-5}
        & \multicolumn{1}{L{0.8cm}}{6} & \multicolumn{1}{L{6cm}}{Late effect of medical and surgical care complication} & \multicolumn{1}{L{2cm}}{63224} & \multicolumn{1}{L{2cm}}{{7.91\%}} \\\cline{2-5}
        & \multicolumn{1}{L{0.8cm}}{7} & \multicolumn{1}{L{6cm}}{Congestive heart failure} & \multicolumn{1}{L{2cm}}{55631} & \multicolumn{1}{L{2cm}}{{6.96\%}} \\\cline{2-5}
        & \multicolumn{1}{L{0.8cm}}{8} & \multicolumn{1}{L{6cm}}{Coronary arteriosclerosis} & \multicolumn{1}{L{2cm}}{53763} & \multicolumn{1}{L{2cm}}{{6.73\%}} \\\cline{2-5}
        & \multicolumn{1}{L{0.8cm}}{9} & \multicolumn{1}{L{6cm}}{Hyperlipidemia} & \multicolumn{1}{L{2cm}}{51459} & \multicolumn{1}{L{2cm}}{{6.44\%}} \\\cline{2-5}
        & \multicolumn{1}{L{0.8cm}}{10} & \multicolumn{1}{L{6cm}}{Diabetes mellitus without complication} & \multicolumn{1}{L{2cm}}{51069} & \multicolumn{1}{L{2cm}}{{6.39\%}} \\\cline{2-5}
  \hline
    \multirow{2}{3cm}{Emergency Room and Inpatient Visit} 
        & \multicolumn{1}{L{0.8cm}}{1} & \multicolumn{1}{L{6cm}}{Essential hypertension} & \multicolumn{1}{L{2cm}}{64914} & \multicolumn{1}{L{2cm}}{30.09\%} \\\cline{2-5}
        & \multicolumn{1}{L{0.8cm}}{2} & \multicolumn{1}{L{6cm}}{Finding related to pregnancy} & \multicolumn{1}{L{2cm}}{43878} & \multicolumn{1}{L{2cm}}{{20.34\%}} \\\cline{2-5}
        & \multicolumn{1}{L{0.8cm}}{3} & \multicolumn{1}{L{6cm}}{Single live birth} & \multicolumn{1}{L{2cm}}{41606} & \multicolumn{1}{L{2cm}}{{19.29\%}} \\\cline{2-5}
        & \multicolumn{1}{L{0.8cm}}{4} & \multicolumn{1}{L{6cm}}{Hyperlipidemia} & \multicolumn{1}{L{2cm}}{39051} & \multicolumn{1}{L{2cm}}{{18.10\%}} \\\cline{2-5}
        & \multicolumn{1}{L{0.8cm}}{5} & \multicolumn{1}{L{6cm}}{Acute renal failure syndrome} & \multicolumn{1}{L{2cm}}{34763} & \multicolumn{1}{L{2cm}}{{16.11\%}} \\\cline{2-5}
        & \multicolumn{1}{L{0.8cm}}{6} & \multicolumn{1}{L{6cm}}{Postpartum finding} & \multicolumn{1}{L{2cm}}{30382} & \multicolumn{1}{L{2cm}}{{14.08\%}} \\\cline{2-5}
        & \multicolumn{1}{L{0.8cm}}{7} & \multicolumn{1}{L{6cm}}{Anemia} & \multicolumn{1}{L{2cm}}{24841} & \multicolumn{1}{L{2cm}}{{11.51\%}} \\\cline{2-5}
        & \multicolumn{1}{L{0.8cm}}{8} & \multicolumn{1}{L{6cm}}{Urinary tract infectious disease} & \multicolumn{1}{L{2cm}}{23831} & \multicolumn{1}{L{2cm}}{{11.05\%}} \\\cline{2-5}
        & \multicolumn{1}{L{0.8cm}}{9} & \multicolumn{1}{L{6cm}}{Chest pain} & \multicolumn{1}{L{2cm}}{22768} & \multicolumn{1}{L{2cm}}{{10.55\%}} \\\cline{2-5}
        & \multicolumn{1}{L{0.8cm}}{10} & \multicolumn{1}{L{6cm}}{Dehydration} & \multicolumn{1}{L{2cm}}{22402} & \multicolumn{1}{L{2cm}}{{10.38\%}} \\
  \hline
    \multirow{2}{3cm}{Home Visit} 
        & \multicolumn{1}{L{0.8cm}}{1} & \multicolumn{1}{L{6cm}}{Essential hypertension} & \multicolumn{1}{L{2cm}}{144} & \multicolumn{1}{L{2cm}}{45.71\%} \\\cline{2-5}
        & \multicolumn{1}{L{0.8cm}}{2} & \multicolumn{1}{L{6cm}}{Malaise} & \multicolumn{1}{L{2cm}}{97} & \multicolumn{1}{L{2cm}}{{30.79\%}} \\\cline{2-5}
        & \multicolumn{1}{L{0.8cm}}{3} & \multicolumn{1}{L{6cm}}{Constipation} & \multicolumn{1}{L{2cm}}{89} & \multicolumn{1}{L{2cm}}{{28.25\%}} \\\cline{2-5}
        & \multicolumn{1}{L{0.8cm}}{4} & \multicolumn{1}{L{6cm}}{Major depression, single episode} & \multicolumn{1}{L{2cm}}{61} & \multicolumn{1}{L{2cm}}{{19.37\%}} \\\cline{2-5}
        & \multicolumn{1}{L{0.8cm}}{5} & \multicolumn{1}{L{6cm}}{Cough} & \multicolumn{1}{L{2cm}}{52} & \multicolumn{1}{L{2cm}}{{16.51\%}} \\\cline{2-5}
        & \multicolumn{1}{L{0.8cm}}{6} & \multicolumn{1}{L{6cm}}{Dementia} & \multicolumn{1}{L{2cm}}{43} & \multicolumn{1}{L{2cm}}{{13.65\%}} \\\cline{2-5}
        & \multicolumn{1}{L{0.8cm}}{7} & \multicolumn{1}{L{6cm}}{Dementia with behavioral disturbance} & \multicolumn{1}{L{2cm}}{42} & \multicolumn{1}{L{2cm}}{{13.33\%}} \\\cline{2-5}
        & \multicolumn{1}{L{0.8cm}}{8} & \multicolumn{1}{L{6cm}}{Disorder due to infection} & \multicolumn{1}{L{2cm}}{42} & \multicolumn{1}{L{2cm}}{{13.33\%}} \\\cline{2-5}
        & \multicolumn{1}{L{0.8cm}}{9} & \multicolumn{1}{L{6cm}}{Slow transit constipation} & \multicolumn{1}{L{2cm}}{41} & \multicolumn{1}{L{2cm}}{{13.02\%}} \\\cline{2-5}
        & \multicolumn{1}{L{0.8cm}}{10} & \multicolumn{1}{L{6cm}}{Hyperlipidemia} & \multicolumn{1}{L{2cm}}{40} & \multicolumn{1}{L{2cm}}{{12.70\%}} \\
  \bottomrule
  \end{tabular}}
\end{table*}

\begin{table*}[hbtp]
\floatconts
  {tab:top_10_most_frequent_concepts_part2}
  {}
  {\begin{tabular}{p{3cm}|p{0.8cm}|p{6cm}|p{2cm}|p{2cm}}
  \toprule
     \multicolumn{1}{p{3cm}}{Visit Type}
      & \multicolumn{1}{p{0.8cm}}{Rank} 
      & \multicolumn{1}{p{6cm}}{Condition Concepts}
      & \multicolumn{1}{p{2cm}}{Patient Count}
      & \multicolumn{1}{p{2cm}}{Percentage of the Visit Type Patients}\\
  \hline
    \multirow{2}{3cm}{Office Visit} 
        & \multicolumn{1}{L{0.8cm}}{1} & \multicolumn{1}{L{6cm}}{Essential hypertension} & \multicolumn{1}{L{2cm}}{62118} & \multicolumn{1}{L{2cm}}{5.83\%} \\\cline{2-5}
        & \multicolumn{1}{L{0.8cm}}{2} & \multicolumn{1}{L{6cm}}{Hyperlipidemia} & \multicolumn{1}{L{2cm}}{36152} & \multicolumn{1}{L{2cm}}{{5.76\%}} \\\cline{2-5}
        & \multicolumn{1}{L{0.8cm}}{3} & \multicolumn{1}{L{6cm}}{Gastroesophageal reflux disease without esophagitis} & \multicolumn{1}{L{2cm}}{26766} & \multicolumn{1}{L{2cm}}{{5.61\%}} \\\cline{2-5}
        & \multicolumn{1}{L{0.8cm}}{4} & \multicolumn{1}{L{6cm}}{Vitamin D deficiency} & \multicolumn{1}{L{2cm}}{26442} & \multicolumn{1}{L{2cm}}{{4.65\%}} \\\cline{2-5}
        & \multicolumn{1}{L{0.8cm}}{5} & \multicolumn{1}{L{6cm}}{Cough} & \multicolumn{1}{L{2cm}}{25727} & \multicolumn{1}{L{2cm}}{{4.48\%}} \\\cline{2-5}
        & \multicolumn{1}{L{0.8cm}}{6} & \multicolumn{1}{L{6cm}}{Pure hypercholesterolemia} & \multicolumn{1}{L{2cm}}{21321} & \multicolumn{1}{L{2cm}}{{4.38\%}} \\\cline{2-5}
        & \multicolumn{1}{L{0.8cm}}{7} & \multicolumn{1}{L{6cm}}{Obesity} & \multicolumn{1}{L{2cm}}{20555} & \multicolumn{1}{L{2cm}}{{4.23\%}} \\\cline{2-5}
        & \multicolumn{1}{L{0.8cm}}{8} & \multicolumn{1}{L{6cm}}{Chronic pain} & \multicolumn{1}{L{2cm}}{20091} & \multicolumn{1}{L{2cm}}{{4.07\%}} \\\cline{2-5}
        & \multicolumn{1}{L{0.8cm}}{9} & \multicolumn{1}{L{6cm}}{Dyspnea} & \multicolumn{1}{L{2cm}}{19387} & \multicolumn{1}{L{2cm}}{{4.01\%}} \\\cline{2-5}
        & \multicolumn{1}{L{0.8cm}}{10} & \multicolumn{1}{L{6cm}}{Fatigue} & \multicolumn{1}{L{2cm}}{18656} & \multicolumn{1}{L{2cm}}{{3.73\%}} \\
  \hline
    \multirow{2}{3cm}{Health examination} 
        & \multicolumn{1}{L{0.8cm}}{1} & \multicolumn{1}{L{6cm}}{Chronic pain} & \multicolumn{1}{L{2cm}}{369} & \multicolumn{1}{L{2cm}}{16.83\%} \\\cline{2-5}
        & \multicolumn{1}{L{0.8cm}}{2} & \multicolumn{1}{L{6cm}}{Shoulder joint pain} & \multicolumn{1}{L{2cm}}{248} & \multicolumn{1}{L{2cm}}{{11.31\%}} \\\cline{2-5}
        & \multicolumn{1}{L{0.8cm}}{3} & \multicolumn{1}{L{6cm}}{Musculoskeletal finding} & \multicolumn{1}{L{2cm}}{172} & \multicolumn{1}{L{2cm}}{{7.84\%}} \\\cline{2-5}
        & \multicolumn{1}{L{0.8cm}}{4} & \multicolumn{1}{L{6cm}}{Low back pain} & \multicolumn{1}{L{2cm}}{158} & \multicolumn{1}{L{2cm}}{{7.20\%}} \\\cline{2-5}
        & \multicolumn{1}{L{0.8cm}}{5} & \multicolumn{1}{L{6cm}}{Interstitial lung disease} & \multicolumn{1}{L{2cm}}{126} & \multicolumn{1}{L{2cm}}{{5.75\%}} \\\cline{2-5}
        & \multicolumn{1}{L{0.8cm}}{6} & \multicolumn{1}{L{6cm}}{Pain in right knee} & \multicolumn{1}{L{2cm}}{124} & \multicolumn{1}{L{2cm}}{{5.65\%}} \\\cline{2-5}
        & \multicolumn{1}{L{0.8cm}}{7} & \multicolumn{1}{L{6cm}}{Pain in left knee} & \multicolumn{1}{L{2cm}}{110} & \multicolumn{1}{L{2cm}}{{5.02\%}} \\\cline{2-5}
        & \multicolumn{1}{L{0.8cm}}{8} & \multicolumn{1}{L{6cm}}{Postoperative state} & \multicolumn{1}{L{2cm}}{103} & \multicolumn{1}{L{2cm}}{{4.70\%}} \\\cline{2-5}
        & \multicolumn{1}{L{0.8cm}}{9} & \multicolumn{1}{L{6cm}}{Difficulty walking} & \multicolumn{1}{L{2cm}}{93} & \multicolumn{1}{L{2cm}}{{4.24\%}} \\\cline{2-5}
        & \multicolumn{1}{L{0.8cm}}{10} & \multicolumn{1}{L{6cm}}{Lumbago with sciatica} & \multicolumn{1}{L{2cm}}{89} & \multicolumn{1}{L{2cm}}{{4.06\%}} \\
  \hline
    \multirow{2}{3cm}{Emergency Room Visit} 
        & \multicolumn{1}{L{0.8cm}}{1} & \multicolumn{1}{L{6cm}}{Abdominal pain} & \multicolumn{1}{L{2cm}}{153538} & \multicolumn{1}{L{2cm}}{13.77\%} \\\cline{2-5}
        & \multicolumn{1}{L{0.8cm}}{2} & \multicolumn{1}{L{6cm}}{Essential hypertension} & \multicolumn{1}{L{2cm}}{123530} & \multicolumn{1}{L{2cm}}{{11.08\%}} \\\cline{2-5}
        & \multicolumn{1}{L{0.8cm}}{3} & \multicolumn{1}{L{6cm}}{Chest pain} & \multicolumn{1}{L{2cm}}{99274} & \multicolumn{1}{L{2cm}}{{8.91\%}} \\\cline{2-5}
        & \multicolumn{1}{L{0.8cm}}{4} & \multicolumn{1}{L{6cm}}{Viral disease} & \multicolumn{1}{L{2cm}}{95402} & \multicolumn{1}{L{2cm}}{{8.56\%}} \\\cline{2-5}
        & \multicolumn{1}{L{0.8cm}}{5} & \multicolumn{1}{L{6cm}}{Headache} & \multicolumn{1}{L{2cm}}{94671} & \multicolumn{1}{L{2cm}}{{8.49\%}} \\\cline{2-5}
        & \multicolumn{1}{L{0.8cm}}{6} & \multicolumn{1}{L{6cm}}{Fever} & \multicolumn{1}{L{2cm}}{92601} & \multicolumn{1}{L{2cm}}{{8.31\%}} \\\cline{2-5}
        & \multicolumn{1}{L{0.8cm}}{7} & \multicolumn{1}{L{6cm}}{Cough} & \multicolumn{1}{L{2cm}}{89758} & \multicolumn{1}{L{2cm}}{{8.05\%}} \\\cline{2-5}
        & \multicolumn{1}{L{0.8cm}}{8} & \multicolumn{1}{L{6cm}}{Acute upper respiratory infection} & \multicolumn{1}{L{2cm}}{88679} & \multicolumn{1}{L{2cm}}{{7.96\%}} \\\cline{2-5}
        & \multicolumn{1}{L{0.8cm}}{9} & \multicolumn{1}{L{6cm}}{Acute pharyngitis} & \multicolumn{1}{L{2cm}}{74356} & \multicolumn{1}{L{2cm}}{{6.67\%}} \\\cline{2-5}
        & \multicolumn{1}{L{0.8cm}}{10} & \multicolumn{1}{L{6cm}}{Asthma} & \multicolumn{1}{L{2cm}}{73301} & \multicolumn{1}{L{2cm}}{{6.58\%}} \\
  \bottomrule
  \end{tabular}}
\end{table*}

\end{document}